\pdfoutput=1
\documentclass[letterpaper, 10 pt, conference]{ieeeconf}  

\IEEEoverridecommandlockouts                              

\usepackage{graphics} 
\usepackage{graphicx} 

\makeatletter
\let\MYcaption\@makecaption
\makeatother

\usepackage[font=scriptsize]{subcaption}

\makeatletter
\let\@makecaption\MYcaption
\makeatother

\usepackage{amsmath} 
\usepackage{amssymb}  
\usepackage{algorithm, algorithmic} 
\usepackage{bm}
\usepackage{xcolor}
\usepackage{dblfloatfix}
\usepackage{multirow}
\usepackage{cite} 
\usepackage[textsize=tiny]{todonotes}

\newcommand{\scriptO}{{O}}
\newcommand{\scripto}{{o}}

\renewcommand{\baselinestretch}{0.98}

\makeatletter
\let\NAT@parse\undefined
\makeatother
\usepackage{hyperref}  

\title{\LARGE \bf
Multi-Robot Planning
on Dynamic Topological Graphs using Mixed-Integer Programming
}

\author{Cora A. Dimmig$^{1,2}$, Kevin C. Wolfe$^{1}$, and Joseph Moore$^{1,2}$
\thanks{$^{1}$Johns Hopkins University Applied Physics Laboratory, Laurel, MD
	20723, USA.}%
\thanks{$^{2}$Department of Mechanical Engineering, Johns Hopkins University, Baltimore, MD 21218, USA.  Email: {\tt\small Cora.Dimmig@jhuapl.edu, Joseph.Moore@jhuapl.edu}}%
}

\begin{document}

\maketitle
\thispagestyle{empty}
\pagestyle{empty}

\begin{abstract}

Planning for multi-robot teams in complex environments is a challenging problem, especially when these teams must coordinate to accomplish a common objective. 
In general, optimal solutions to these planning problems are computationally intractable, since the decision space grows exponentially with the number of robots. In this paper, we present a novel approach for multi-robot planning on topological graphs using mixed-integer programming. Central to our approach is the notion of a dynamic topological graph, where edge weights vary dynamically based on the locations of the robots in the graph. We construct this graph using the critical features of the planning problem and the relationships between robots; we then leverage mixed-integer programming to minimize a shared cost that depends on the paths of all robots through the graph. 
To improve computational tractability, we formulated our optimization problem with a fully convex relaxation and designed our decision space around eliminating the exponential dependence on the number of robots. 
We test our approach on a multi-robot 
reconnaissance scenario, where robots must coordinate to minimize detectability and maximize safety while gathering information. 
We demonstrate that our approach is able to scale to a series of representative scenarios 
and is capable of computing optimal 
coordinated strategic behaviors for autonomous multi-robot teams in seconds.

\end{abstract}

\section{INTRODUCTION}

Achieving unified coordination in complex real-world environments is a fundamental challenge for multi-robot systems. The robots must often achieve dynamic task allocation under geometric and temporal constraints. The problem is further complicated if teaming is required in the presence of uncertainty, such as might arise from inaccurate motion models or imperfect communication. In many circumstances, the combinatorial complexity of these multi-robot planning problems leads to computational intractability. This is especially true as the number of robots increases. 

In this paper, we present a novel approach for multi-robot planning on topological graphs using mixed-integer programming (MIP). In particular, our approach leverages the notion of a \emph{dynamic} topological graph, where the edge weights vary with the state of the robot team. To construct our graphs, we use a problem-specific embedding that captures the important features of the planning problem. We then use MIP to formulate a compact optimization problem which calculates paths for multi-robot teams that collaborate to minimize a shared cost function. We formulated our optimization problem with a fully convex relaxation to be able to compute solutions to challenging real world scenarios in seconds, and overcome the common computational limitations of MIP.
These rapid solve times have the potential to enable frequent online re-planning in evolving conditions.
Additionally, by using MIP, our results are interpretable and extensible, and we can calculate a solution with guaranteed optimality. 

\begin{figure}[t]
	\vspace*{2mm}
	\centering
	\includegraphics[trim={0cm 1.2cm 0cm 1.2cm},clip, width=1.0\columnwidth]{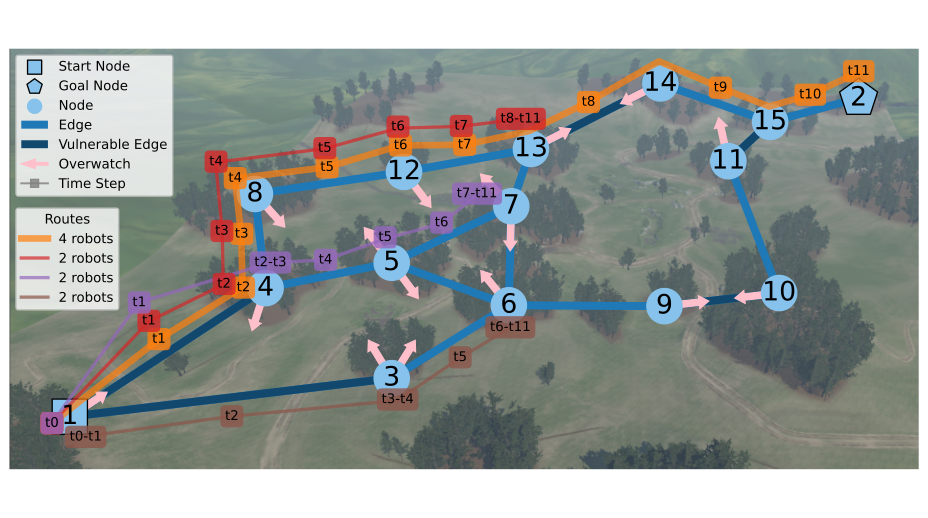}
	\caption{A depiction of our proposed dynamic topological graph applied to a reconnaissance problem. 
    The nodes are in forested regions of cover. Ten robot scouts start at node 1 with the goal of at least one robot reaching node 2, across the meadow. 
    The edges of the graph have cost for transitioning between nodes that is a function of their distance, detectability, and vulnerability.  
    This cost can be reduced by moving in teams and by providing overwatch, where robots at a particular node oversee the movement of robots along a corresponding edge to help mitigate some of the risk of traversing. Overwatch opportunities are indicated with an arrow from the overwatch node pointing toward the edge that can be monitored. 
    The robot team routes through the graph, solved for with our proposed method, represent a solution to this problem.
	}
	\label{fig:built_side_problem}
	\vspace*{-3mm}
\end{figure}

The dynamic topological graph structure we propose also facilitates the transfer of problem specific elements to new application domains. For this reason, we believe our approach has the potential to be generally applicable to a broad set of complex scenarios.
In this work, we demonstrate our approach on sample reconnaissance test cases, e.g. Fig.~\ref{fig:built_side_problem}, where teams of autonomous robots coordinate to maneuver through complex environments while minimizing detection. 
%

\section{RELATED WORK}

Multi-robot coordination \cite{Yan2013, Verma2021} and cooperative multi-agent planning \cite{Torreno2017} 
are computationally challenging spaces. 

\subsection{General Methods}

For multi-agent path finding, Conflict Based Search (CBS) methods \cite{Sharon2015} plan optimal paths for all agents and then resolve conflicts in the paths,  
ultimately planning to avoid interactions between agents. 
Multi-agent task allocation approaches often consider robots working independently toward a shared objective, such as with auction-based methods \cite{Lagoudakis2004} and game theoretic approaches \cite{Park2021}. 
We look to encourage explicit cooperation between robots towards a goal, which requires more directly considering their interactions. 

Multi-agent games require cooperation between agents to be considered, though often necessitate decomposing the problem into smaller local-games, as in \cite{Shishika2018}, 
which hinders finding a fully optimal solution to the overarching problem. 

\subsection{Machine Learning}

Recently, techniques utilizing Machine Learning (ML) have emerged,
commonly using Graph Neural Networks \cite{Zhou2020, Kortvelesy2021}, such as for 
scheduling problems \cite{Wang2022}, 
handling agent interactions \cite{Liu2020}, and considering coordination in uncertain and adversarial environments \cite{Zhou2021}.
One major challenge with ML techniques is generalization to scenarios that were not observed during training. Some recent work with Reinforcement Learning (RL) have shown progress in this space \cite{Paul2022, Almasan2022}. However, these approaches either train and test in similar data distributions or report degraded performance in new scenarios. Additionally, optimality cannot be guaranteed. 

Some work has been done with Multi-agent Markov Decision Processes toward RL algorithms with optimality guarantees \cite{Qu2019}. However, categorically, ML approaches have not been able to guarantee optimality, and while they aid in overcoming computational limitations of conventional approaches, they suffer from low interpretability. 

\subsection{Mixed-Integer Programming}

A Mixed-Integer Programming (MIP) problem is an optimization problem in which decision variables may be real or integer valued. 
This enables efficient, interpretable problem formulations that can be solved to optimality. 

In \cite{Yu2016}, complete algorithms for multi-robot path planning are reported with heuristics to improve computational performance.  
However, the authors do not investigate the complexity of explicit collaboration between robots.
Similarly, MIP is used in recharging or timed delivery scenarios, as in \cite{Mathew2013, Kamra2015}, 
which considers planning for delivery robots around task robots with fixed locations or known paths. 
In our proposed approach we simultaneously optimize paths of collaborating robots. Explicit collaboration in our method introduces the dynamic element to our topological graphs and enables devising optimal plans across the entire team.

Temporal Logic has been applied to compactly define complex task specifications on a fixed graph, e.g. \cite{Leahy2022,Sahin2020}. In contrast, our approach focuses on an embedding that encourages emergent behaviors through dynamic edge weights and a state dependent graph. However, similar to our approach, the authors in \cite{Leahy2022,Sahin2020} remove the focus on tracking paths for each individual agent, in favor of tracking the number of robots satisfying a condition. 

MIP is NP-complete, which can inherently lead to long solve times. 
Many MIP applications can accommodate longer solve times, such as for 
scheduling and task allocation \cite{Koes2005} and communication provisioning \cite{Flushing2017}. 
In \cite{Asfora2020}, the authors explore intercepting a moving target, though their approach requires computation offline for an optimal solution. 
In our application space, these long solve times would not be practical.
Fortunately, modern solving methods, e.g. the Branch-and-Bound/Cut algorithms we used in this work from Gurobi \cite{GurobiOptimization2023}, and efficient problem formulations, e.g. as developed in \cite{Marcucci2021}, have helped to drastically improve solve times. 
In the work reported herein, we have used these results and formulated our problem with a fully convex relaxation, which expedites solve times with Branch-and-Bound methods. We demonstrate solving large scale, complex problems entirely to optimality in seconds, enabling online applications and re-planning as operational conditions evolve. 

\section{PROBLEM STATEMENT}
\label{sec:problem_statement}

Our objective is to plan tactical maneuvers for a multi-robot team that minimize detection, maximize safe navigation between regions of cover, and give rise to tactical behaviors such as ``bounding overwatch.'' Bounding overwatch is where teams alternate movement across dangerous areas to oversee the other teams movement and mitigate risk.
We consider three foundational constructs for reducing the risk of traversing a path:
\begin{enumerate}
	\item \textbf{Overwatch opportunities}: When one team is at a vantage point to oversee the movement of another team, they are said to be providing overwatch, which reduces the risk of movement for the traversing team.
	\item \textbf{Formations in areas of high vulnerability}: When an area to traverse is particularly dangerous, moving with a greater number of robots allows moving in a formation to increase awareness. 
	\item \textbf{General teaming}: Moving in teams offers a higher level of protection to the robots.
\end{enumerate}
We will refer to these constructs as overwatch, vulnerability, and teaming, respectively.

To achieve risk-sensitive coordinated reconnaissance 
with a multi-robot team, we propose representing this problem as a topological graph with dynamic edge weights. 
Nodes in the graph represent regions of cover and/or essential locations, while edges represent traversable paths between the nodes. 
Our three foundational constructs set the weights on the edges in our graphs dynamically, based on the locations of the robots in the graph. 
The base edge weight is a function of detectability and distance. Then if a particular overwatch position is occupied, corresponding edges will have reduced weights. 
The cost of the edge is increased if a desired number of robots for a vulnerable edge is not met and edge cost is reduced based on the number of robots traversing the edge as an incentive for moving as a team.
Fig.~\ref{fig:built_side_problem} depicts a sample problem using a dynamic topological graph.

\subsection{Combinatorial Considerations}

For a total number of robots/agents $n_A$ and total number of locations $n_L$ (nodes, $n_V$, and edges, $n_E$), all possible states of robots at locations in a graph for a particular time step grows exponentially as follows.
\begin{align}
{n_L}^{n_A}
\label{eq:combinatorics}
\end{align}
Thus, our problem would quickly become computationally intractable for methods that seek to 
apply dynamic programming in this discrete state space. Techniques that rely on using function approximation to overcome the curse of dimensionality would likely suffer from poor generalizability and explainability.

To improve computational tractability, 
we developed a compact MIP 
formulation to solve the multi-robot planning problem. 
In our formulation, we were able to drastically reduce the size of the decision space. 
When considering $n_\scriptO$ overwatch opportunities and $n_T$ time steps, our total number of variables scales by (\ref{eq:vars_scaling}), which is linear in the parameters. 
\begin{align}
n_T (1 + n_L + 2 n_E + n_\scriptO) \label{eq:vars_scaling}
\end{align}
This is comprised of $n_T(1 + n_E)$ binary, $n_T n_L$ integer, and $n_T (n_E + n_\scriptO)$ continuous variables.
Additionally, we were able to remove the dependence on the number of robots from our decision space. In this work, we assume a homogeneous team of robots, though this approach could scale by the number of different types of robots for heterogeneous teams. 
In this paper, we demonstrate that our MIP formulation stays tractable for larger graph sizes. Furthermore, our approach is easily explainable for interpreting the results and applicable to new problem spaces.

\section{MIXED-INTEGER PROGRAMMING APPROACH}

The objective of our mixed-integer program is to position a group of robots at a set of target locations within a specified time horizon while minimizing time and cost due to traversing edges. 
In our scenario, the costs of the edges represent the risk associated with being detected or disabled while traversing. 
Using our foundational constructs from Section \ref{sec:problem_statement}, certain nodes will provide overwatch of particular edges; this construct reduces the cost of traversing that edge since traversal risk has been reduced. 
We consider reduced risk/cost when multiple robots are traversing a particular edge, and, for particularly vulnerable edges, we further incentivize more robots to move together. 

Table~\ref{tab:parameters} lists the parameters we use to formulate the multi-robot reconnaissance problem
and construct our decision variables. 
Table \ref{tab:decision_vars} shows each group of decision variables, their type (integer, binary, or continuous), lower bounds (LB), upper bounds (UB), and descriptions.

\begin{table}
	\centering
    \vspace*{2mm}
	\caption{MIP Parameters}
	\vspace*{-2mm}
    \label{tab:parameters}
	\begin{center}
		\renewcommand{\arraystretch}{1.3}
		\begin{tabular}{ c | c | p{4.9cm} }
			\textbf{Category} & \textbf{Var} & \textbf{Description} \\
			\hline
			\hline
			\multirow{8}{4.5em}{Problem Size}
			& $n_A$ & Number of agents/robots \\
			& $n_T$ & Number of time steps in the time horizon \\ 
			& $n_\scriptO$ & Number of overwatch opportunities \\
			& $n_E$ & Number of edges, both directions \\
			& $n_V$ & Number of nodes/vertices \\
			& $n_L$ & Number of locations ($n_E + n_V$) \\
			& $n_S$ & Number of start locations \\
			& $n_G$ & Number of goal locations \\
   			\hline
			\multirow{8}{4.5em}{Scenario Variables} 
            & $E$ & Set of edges $e$ \\
            & $V$ & Set of nodes/vertices $v$ \\
            & $L$ & Set of locations $l$ consisting of edges and vertices, $E \cup V$ \\
            & $S$ & Set of start locations $s$, $S \subseteq L$ \\
            & $G$ & Set of goal locations $g$, $G \subseteq L$ \\
            & $\scriptO$ & Set of overwatch opportunities $\scripto$, $(v_i, e_j)$ where node $v_i$ can overwatch edge $e_j$ \\
			\hline
			\multirow{3}{4.5em}{Problem Parameters} 
			& $t$ & Time step from $1$ to $n_T$ \\
			& $n_{s}$ & Number of robots at start location $s \in S$ \\
			& $n_{g}$ & Number of robots at goal location $g \in G$ \\
			\hline
			\multirow{4}{4.5em}{Cost of Traversing} 
			& $w_e$ & Base cost to traverse edge $e \in E$ \\
			& $a_e$ & Minimum desired number of robots on $e$ \\ 
			& $m_e$ & Additional cost for robots on $e$ before $a_e$ \\
			& $r_e$ & Cost reduction on $e$ for robots over $a_e$ \\
			\hline
			\multirow{3}{4.5em}{Cost of Overwatch} 
			& $\omega_\scripto$ & Benefit of full overwatch for $\scripto~\in~\scriptO$ \\
			& $\alpha_\scripto$ & Number of robots for full overwatch for $\scripto$ \\ 
			& $\gamma_\scripto$ & Reward for overwatch robots over $\alpha_\scripto$ for~$\scripto$ \\ 
		\end{tabular}
	\end{center}
\end{table}
\begin{table}[tbh]
    \vspace*{-2mm}
	\centering
	\caption{MIP Decision Variables (At Time $t$)}
    \vspace*{-2mm}
	\label{tab:decision_vars}
	\begin{center}
		\renewcommand{\arraystretch}{1.3}
		\begin{tabular}{ c | c | c | c | p{3.65cm} }
			\textbf{Var} & \textbf{Type} & \textbf{LB} & \textbf{UB} & \textbf{Description} \\
			\hline
			\hline
			$p_{l,t}$ & Integer & 0 & $n_A$ & Number of robots at location $l$ \\ 
			$\phi_{e,t}$ & Binary & 0 & 1 & Whether robots are on edge $e$ \\ 
			$\psi_{t}$ & Binary & 0 & 1 & Whether robots have moved \\ 
			$C_{W_{e,t}}$ & Cont. & 0 & $\infty$ & Cost of traversing edge $e$ \\ 
			$C_{\Omega_{\scripto,t}}$ & Cont. & $-\infty$ & 0 & Cost of overwatch opportunity $\scripto$ \\ 
		\end{tabular}
	\end{center}
    \vspace*{-6mm}
\end{table}

\subsection{Cost Function}

Our goal is to minimize the overall cost of traversing and the total time to reach a set of target locations. 

\subsubsection{Cost of Traversing}

\begin{figure}[tbh]
    \vspace*{3mm}
	\centering
	\begin{subfigure}[t]{0.49\columnwidth}
		\footnotesize
		\centering
		\includegraphics[width=\textwidth]{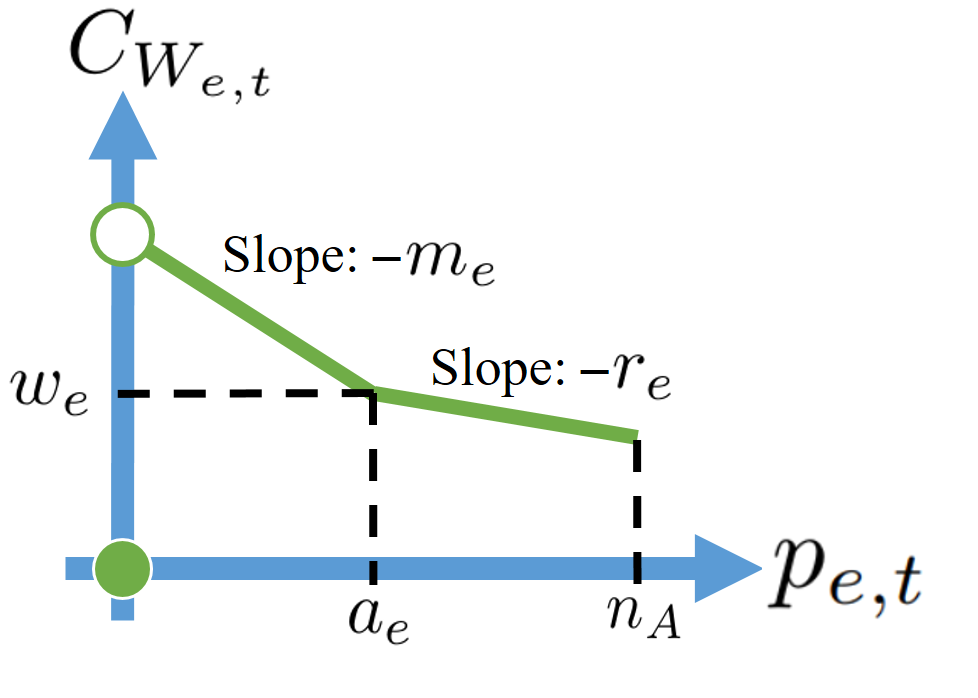}
		\scriptsize
		\caption{Cost of traversing edge~$e$ at time~$t$ versus number of robots on the edge}
		\label{fig:cost_of_traversing}
	\end{subfigure}
	\hfill
	\begin{subfigure}[t]{0.49\columnwidth}
		\footnotesize
		\centering
		\includegraphics[width=\textwidth]{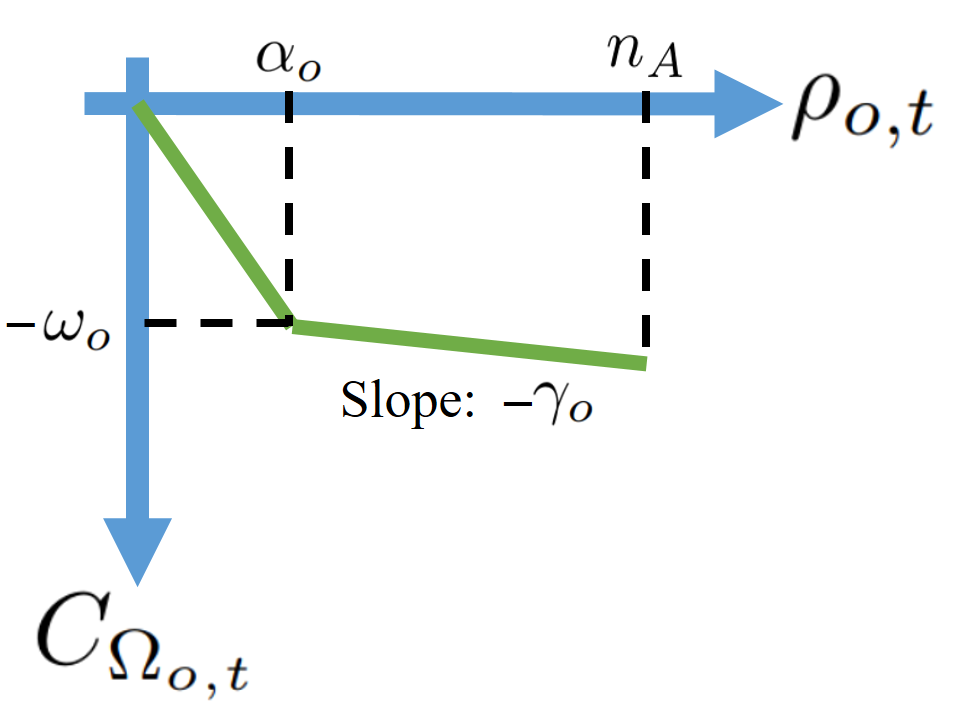}
        \scriptsize
		\caption{Cost of overwatch opportunity~$\scripto$ at time~$t$ versus number of overwatch robots}
		\label{fig:cost_of_overwatch}
	\end{subfigure}
	\caption{Piecewise-Linear Cost Functions}
	\label{fig:pwl_costs}
    \vspace*{-4mm}
\end{figure}

Using the parameters from Table~\ref{tab:parameters}, we consider a base cost to traverse an edge $e$ as a positive $w_e$. We encode vulnerability information for that edge by specifying a minimum desired number of robots $a_e$ on the edge and additional positive cost $m_e$ for each robot until the minimum is met.
More robots on a vulnerable edge would enable moving in a formation for greater awareness. 
As an incentive for further teaming, for each robot over the minimum desired, an additional reduction can be specified as a positive $r_e$. Thus, as depicted in Fig.~\ref{fig:cost_of_traversing}, the piecewise-linear cost of traversing an edge at a particular time, $C_{W_{e,t}}$, 
can be expressed in terms of the number of robots on the edge at that time, $p_{e, t}$, as follows.
\begin{align}
C_{W_{e,t}} = \begin{cases}
0, & p_{e, t} = 0 \\
w_e + m_e (a_e - p_{e, t}), & 0 < p_{e, t} \leq a_e \\
w_e - r_e(p_{e, t} - a_e), & a_e \leq p_{e, t} \leq n_A
\end{cases}
\label{eq:pwl_traversing}
\end{align}

When our integrality constraints are relaxed, this cost is not convex due to the zero point. 
We can restate this cost
using perspective functions, as described in \cite{Marcucci2021}, and a variable for whether an edge is used, $\phi_{e,t}$, for the relaxation to be convex. This allows the first two cases to be combined. 
\begin{align}
C_{W_{e,t}} = \begin{cases}
-m_e p_{e, t} + (w_e + m_e a_e) \phi_{e,t}, & 0 \leq p_{e, t} \leq a_e \\
w_e - r_e(p_{e, t} - a_e), & a_e \leq p_{e, t} \leq n_A
\end{cases}
\label{eq:pwl_traversing_convex}
\end{align}

When $p_{e, t} = 0$ then $\phi_{e,t} = 0$ by definition, and this cost is zero. Otherwise, for any $p_{e, t} > 0$ then $\phi_{e,t} = 1$ and the first case in (\ref{eq:pwl_traversing_convex}) is identical to the second case in (\ref{eq:pwl_traversing}). 

The cost in (\ref{eq:pwl_traversing_convex}) is convex when we select $m_e \geq r_e$ and relax our integrality constraints. Due to this convexity,
we can equivalently express this cost in our optimization problem with a linear term in the cost function 
that scales with our decision variables $C_{W_{e,t}}$ and two linear constraints. 
\begin{align}
C_{W_{e,t}} &\geq - m_e p_{e, t} + (w_e + m_e a_e) \phi_{e,t} \label{eq:cost:trav1} \\
C_{W_{e,t}} &\geq - r_e p_{e, t} + (w_e + r_e a_e) \phi_{e,t} \label{eq:cost:trav2} 
\end{align}

Since we are optimizing for minimum cost, $C_{W_{e,t}}$ will be tight to the piecewise-linear cost function (\ref{eq:pwl_traversing_convex}). 

\subsubsection{Cost of Overwatch}

For each edge with an overwatch opportunity, we consider having overwatch while traversing to offer a cost reduction. When robots are traversing a corresponding edge, any number of robots at the overwatch node results in overwatch, but more robots may provide a greater reward. 
Using the parameters from Table~\ref{tab:parameters}, we specify a positive $\omega_\scripto$ as the benefit of overwatch opportunity~$\scripto$, $\alpha_\scripto$~is the number of robots needed for full overwatch (i.e. to receive the full reward), and a positive $\gamma_\scripto$ is a further reward for additional robots past $\alpha_\scripto$ (i.e. a further teaming incentive). We consider $\rho_{\scripto,t}$ to be the number of overwatch robots for opportunity $\scripto$ at time $t$. Thus, as depicted in Fig.~\ref{fig:cost_of_overwatch}, for a particular overwatch opportunity the ``cost'' is $C_{\Omega_{\scripto,t}}$. 
\begin{align}
C_{\Omega_{\scripto,t}} = \begin{cases}
-\frac{\omega_\scripto}{\alpha_\scripto} \rho_{\scripto,t}, & 0 \leq \rho_{\scripto,t} \leq \alpha_\scripto \\
-\omega_\scripto - \gamma_\scripto (\rho_{\scripto,t} - \alpha_\scripto), & \alpha_\scripto \leq \rho_{\scripto,t} \leq n_A
\end{cases}
\label{eq:pwl_overwatch}
\end{align}

This ``cost'' will always be negative since it is rewarding overwatch. Similar to the cost for traversing, since this piecewise-linear cost is convex when integrality constraints are relaxed and $\omega_\scripto / \alpha_\scripto \geq \gamma_\scripto$, we can express the cost, $C_{\Omega_{\scripto,t}}$, linearly in our cost function 
with 
two linear constraints. 
Additionally, we can remove the overwatch robot variables, $\rho_{\scripto,t}$, in these constraints. 
For a particular $\scripto \in \scriptO$, we consider the number of robots providing overwatch to be equal to the number of robots at the node, $v_i$, if there are robots on the edge, $e_j$.  
We set $\rho_{\scripto,t} = p_{v_i, t}$, resulting in the following constraints.
\begin{align}
C_{\Omega_{\scripto,t}} &\geq - \frac{\omega_\scripto}{\alpha_\scripto} p_{v_i, t} \label{eq:cost:overwatch1} \\
C_{\Omega_{\scripto,t}} &\geq - \omega_\scripto - \gamma_\scripto (p_{v_i, t} - \alpha_\scripto) 
\label{eq:cost:overwatch2}
\end{align}
To ensure there are robots traversing the corresponding edge~$e_j$, we add another constraint dependent on $p_{e_j, t}$. 
\begin{align}
C_{\Omega_{\scripto,t}} \geq - \frac{\omega_\scripto}{\alpha_\scripto} n_A p_{e_j, t} 
\label{eq:cost:overwatch3}
\end{align}
This constraint 
has a steeper slope (due to scaling by~$n_A$) than
(\ref{eq:cost:overwatch1}) and (\ref{eq:cost:overwatch2})
when there are robots on the edge, i.e. when $p_{e_j, t}~>~0$, to assure it is the least restrictive. 
For $p_{v_i, t}~>~0$, 
(\ref{eq:cost:overwatch3})
is the most restrictive overwatch constraint when $p_{e_j, t}~=~0$, forcing the cost $C_{\Omega_{\scripto,t}}$ to be zero, since there is not an overwatch benefit when there are not robots traversing the edge. Similarly, when $p_{v_i} = 0$, $C_{\Omega_{\scripto,t}} = 0$ since there are not robots providing overwatch. 

\subsubsection{Cost of Time}

The final cost, $C_T$, is for minimizing the time to achieve the goal.
We formulate a cost that scales with time for each time step robots are moving and thus rewards achieving the goal as quickly as possible. This cost uses the binary decision variables $\psi_{t}$ that represent whether robots have moved at time $t$. 
\begin{align}
C_T = \sum_{t = 1}^{n_T} t \psi_t
\end{align}

\subsubsection{Overall Cost}

The overall cost to minimize 
can be expressed as follows.
\begin{align}
C = C_T + \sum_{t = 1}^{n_T} \bigg( \sum_{e \in E} C_{W_{e,t}} + \sum_{\scripto \in \scriptO} C_{\Omega_{\scripto,t}}\bigg) \label{eq:overall_cost}
\end{align}
These terms could be weighted depending on the priority of minimizing traversing cost versus minimizing time. 

\subsection{Constraints}

We add constraints to set support variables used in our cost functions and restrict movement to the 
topological graph.

\subsubsection{Edge Used Variables}

The following constraint sets
the variables tracking if an edge is used, $\phi_{e,t}$, to true if there are robots on that edge, and false otherwise. This constraint assumes that setting this variable to true will never reduce the cost. 
For each time step $t$ and edge $e$, we bound $\phi_{e,t}$.
\begin{align}
\phi_{e,t} \geq \frac{1}{n_A} p_{e,t} \label{eq:constr:edge_used}
\end{align}

\subsubsection{Time Tracking Variables}

We add binary variables, $\psi_t$, to track if there are robots on the edges of the graph. Assuming the edges all have weight, 
we sum the number of robots on the edges for each time step to track whether robots are still moving since robots cannot go instantaneously between nodes. 
This constraint sets $\psi_t = 1$ if the sum is nonzero and $\psi_t = 0$ otherwise, since this variable contributes to increasing the cost.
For each time step $t$, we bound $\psi_t$. 
\begin{align}
\psi_t \geq \frac{1}{n_A} \sum_{e \in E} p_{e,t} \label{eq:constr:time_vars}
\end{align} 

\subsubsection{Start Locations}

We add constraints for the start locations of each robot. For each start location $s \in S$, with $n_{s}$ robots at that location, we add the following constraint. 
\begin{align}
p_{s, 1} = n_{s} \label{eq:constr:start}
\end{align}

\subsubsection{Goal Locations}

For each goal location $g \in G$ which requires at least $n_{g}$ robots, we add the following constraint.
\begin{align}
p_{g, n_T} \geq n_{g} \label{eq:constr:goal}
\end{align}

\subsubsection{Maximum Robots}

To ensure that only the maximum number of robots can exist across all locations at a particular time, 
for each time $t$, we add the following constraint.
\begin{align}
\sum_{l \in L} p_{l,t} = n_A \label{eq:constr:max_robots}
\end{align}

\subsubsection{Sequential Flow}

For each node, the number of robots in the node and flowing into the node must be equal to the number of robots in the node and flowing out of the node in the next time step.
Thus, for each time $t \in [2, n_T]$ and node~$v_j$, we add the following constraint.
\begin{align}
\sum_{\substack{l_{ij} = (v_i, v_j) \in L}} p_{l_{ij},t-1} = \sum_{\substack{l_{ji} = (v_j, v_i) \in L}} p_{l_{ji},t}  \label{eq:constr:sequential}
\end{align}
The first sum 
considers all locations of the form $l_{ij} = (v_i, v_j)$ and the second sum considers all locations of the form $l_{ji} = (v_j, v_i)$ for a fixed node $v_j$. 
Both sets of locations include $l_{jj}$ since all nodes have self-loops.  
This constraint assumes a robot would never stay on an edge (due to the cost). This is enforced by the cost function as long as the cost of all edges is greater than zero and overwatch is never provided from an edge. 
This constraint allows robots to move from one edge to the next without stopping at the node.

\subsection{Optimization Problem}
\label{sec:optimization_problem}

Combining our objective function and constraints from the previous sections, our overall MIP optimization problem is expressed in Table~\ref{tab:optimization_problem}.

\begin{table}[b]
\centering
    \vspace*{-2mm}
	\caption{MIP Optimization Problem}
    \vspace*{-4mm}
	\label{tab:optimization_problem}
	\begin{center}
		\renewcommand{\arraystretch}{2.0}
        \resizebox{\columnwidth}{!}{%
		\begin{tabular}{ p{0.01cm} p{5.54cm} p{1.31cm} | c }
			\multicolumn{3}{c|}{\textbf{Optimization Problem}} & \textbf{Eq.} \\
			\hline
			\hline
			\multicolumn{3}{l|}{$\min~C_T + \sum\limits_{t = 1}^{n_T} \bigg( \sum\limits_{e \in E} C_{W_{e,t}} + \sum\limits_{\scripto \in \scriptO} C_{\Omega_{\scripto,t}}\bigg)$ subject to} & (\ref{eq:overall_cost})\\
            \hline
            \multirow{5}{*}{\rotatebox[origin=c]{90}{Cost Constraints~}} 
            & $C_{W_{e,t}} \geq - m_e p_{e, t} + (w_e + m_e a_e) \phi_{e,t},$ &$\forall e, t$ & (\ref{eq:cost:trav1})\\
            & $C_{W_{e,t}} \geq - r_e p_{e, t} + (w_e + r_e a_e) \phi_{e,t},$ &$\forall e, t$ & (\ref{eq:cost:trav2})\\
            & $C_{\Omega_{\scripto,t}} \geq - \frac{\omega_\scripto}{\alpha_\scripto} p_{v_i, t},$ &$\forall \scripto, t $ & (\ref{eq:cost:overwatch1})\\
            & $C_{\Omega_{\scripto,t}} \geq - \omega_\scripto - \gamma_\scripto (p_{v_i, t} - \alpha_\scripto),$ &$\forall \scripto, t $ & (\ref{eq:cost:overwatch2})\\
            & $C_{\Omega_{\scripto,t}} \geq - \frac{\omega_\scripto}{\alpha_\scripto} n_A p_{e_j, t},$ &$\forall \scripto, t $ & (\ref{eq:cost:overwatch3})\\
            \hline
            \multirow{6}{*}{\rotatebox[origin=c]{90}{Constraints\quad\quad\quad}} 
            & $\phi_{e,t} \geq \frac{1}{n_A} p_{e,t},$ & $\forall e, t $ & (\ref{eq:constr:edge_used}) \\
            & $\psi_t \geq \frac{1}{n_A} \sum\limits_{e \in E} p_{e,t},$ &$\forall t $ & (\ref{eq:constr:time_vars}) \\
            & $p_{s, 1} = n_{s},$ &$\forall s $ & (\ref{eq:constr:start})\\
            & $p_{g, n_T} \geq n_{g},$ &$\forall g $ & (\ref{eq:constr:goal})\\
            & $\sum\limits_{l \in L} p_{l,t} = n_A,$ &$\forall t$ & (\ref{eq:constr:max_robots})\\
            & $\sum\limits_{l_{ij} = (v_i, v_j) \in L} p_{l_{ij},t-1} = \sum\limits_{l_{ji} = (v_j, v_i) \in L} p_{l_{ji},t},$ &$\forall v_j, \newline t \in [2, n_T]$ & (\ref{eq:constr:sequential}) \\
		\end{tabular}%
        }
	\end{center}
    \vspace*{-3mm}
\end{table}

By capturing the piecewise-linear cost structures with linear constraints due to the convexity of the costs (with relaxed integrality constraints), we can use mixed-integer linear programming to solve our problem since we have a linear objective function with linear constraints. 

\subsection{Cost Formulation Considerations}

We assume a graph without negative cycles.
Overwatch for one edge can come from multiple nodes, but the weight of the edge (resulting from the cost of traversing, benefit of overwatch, and cost reductions from vulnerability/teaming) cannot reduce to 0 or below. 
The formulation proposed herein allows robots at one node to overwatch robots traversing multiple edges. 
Intuitively, for an overwatch position to be used, the cost reduction from overwatch needs to be more than the cost to get to the overwatch position. 
Otherwise, 
the robots will stay together for teaming benefits.

While our solutions are always guaranteed to be optimal, the optimal solution is not guaranteed to be unique. Small problems with simple numbers can result in 
many equivalent cost solutions and cause computation time to increase. 
For example, with teaming, without any other factors such as vulnerability and overwatch, if the $r_e$ cost reductions are equivalent on all edges and there are multiple routes being taken, more robots will go on the longest route, which is logical from a risk mitigation perspective. If route lengths are equal, multiple optimal solutions will arise. More realistic values for the cost parameters will diminish these challenges. In the future, we plan to use reinforcement learning to generate appropriate weights for operational scenarios.

\section{COMPUTATIONAL RESULTS AND DISCUSSION}

For scenarios of interest, we create topological graphs and solve our MIP optimization problem from Table~\ref{tab:optimization_problem} using Gurobi \cite{GurobiOptimization2023}. We then process this solution through an assignment routine to determine paths for each robot through the graph. In this section, we present a few sample scenarios. 

\subsection{Illustrative Example}

\begin{figure}
    \vspace*{2mm}
	\centering
	\begin{subfigure}[t]{0.65\columnwidth}
		\centering
		\includegraphics[width=\textwidth]{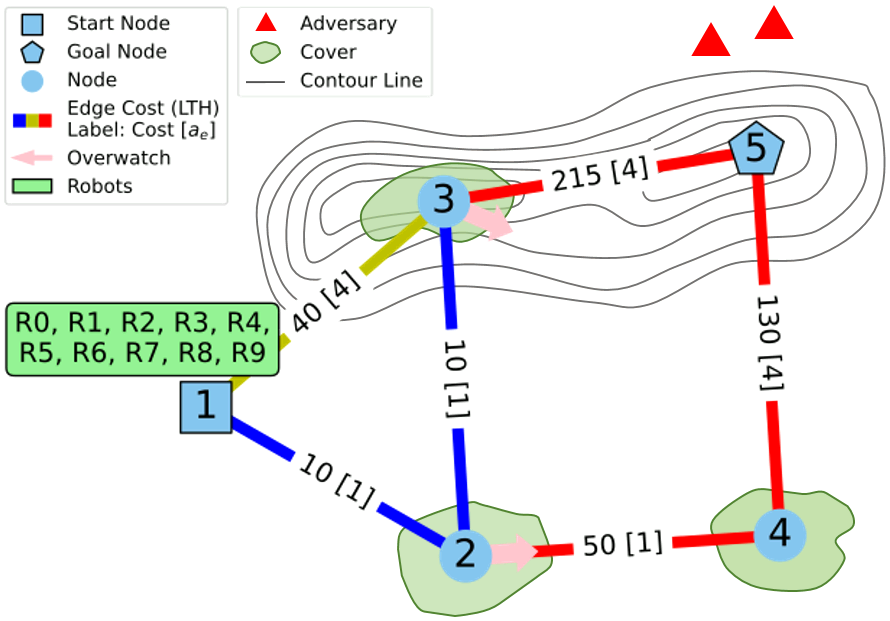}
		\caption{Time = 0, Total Cost = 0}
		\label{fig:toy_problem_solution_0}
	\end{subfigure}%
	
	\begin{subfigure}[t]{0.5\columnwidth}
		\centering
		\includegraphics[trim={1.7cm 1.9cm 0.9cm 0.3cm},clip, width=\textwidth]{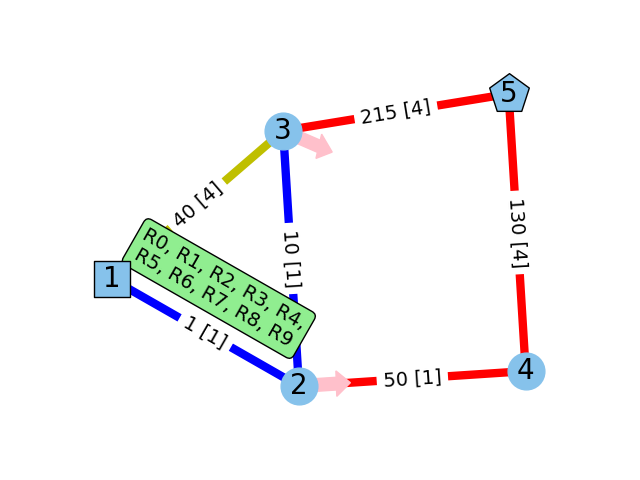}
		\caption{Time = 1, Total Cost = 11}
		\label{fig:toy_problem_solution_1}
	\end{subfigure}%
	\begin{subfigure}[t]{0.5\columnwidth}
		\centering
		\includegraphics[trim={1.7cm 1.9cm 0.9cm 0.3cm},clip, width=\textwidth]{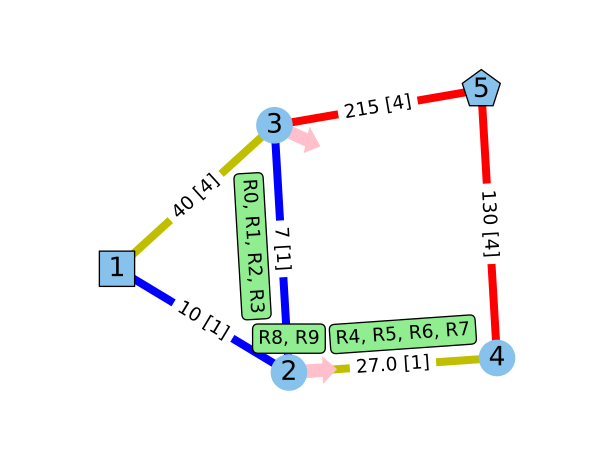}
		\caption{Time = 2, Total Cost = 65}
		\label{fig:toy_problem_solution_2}
	\end{subfigure}
	
	\begin{subfigure}[t]{0.5\columnwidth}
		\centering
		\includegraphics[trim={1.7cm 1.9cm 0.9cm 0.3cm},clip, width=\textwidth]{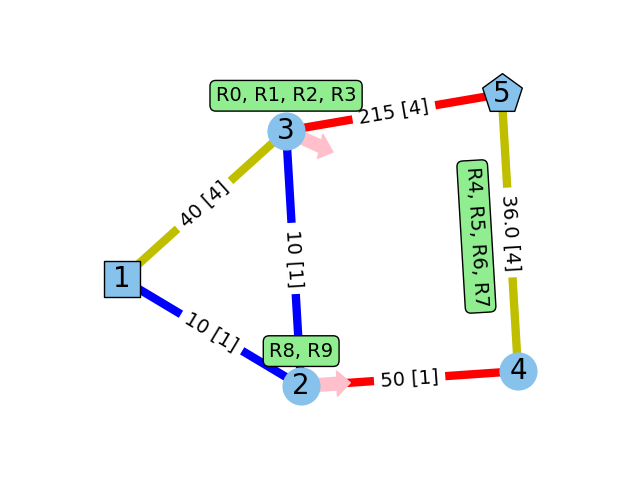}
		\caption{Time = 3, Total Cost = 131}
		\label{fig:toy_problem_solution_3}
	\end{subfigure}
	\begin{subfigure}[t]{0.5\columnwidth}
		\centering
		\includegraphics[trim={1.7cm 1.9cm 0.9cm 0.3cm},clip, width=\textwidth]{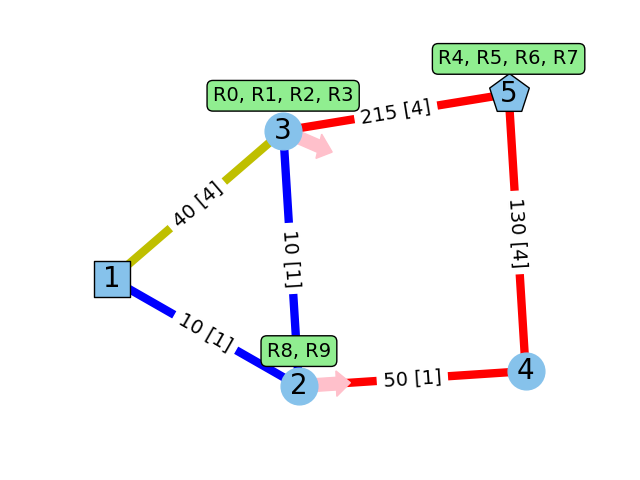}
		\caption{Time = 4, Total Cost = 131}
		\label{fig:toy_problem_solution_4}
	\end{subfigure}%
	\caption{MIP problem solution to an illustrative multi-robot reconnaissance scenario, sketched in 
    (a).
    Ten robots start at node 1 with a goal of at least one robot reaching node 5 to observe the adversary units. The light green regions represent areas of cover. The color of the edges indicate the threat level for transitioning between nodes from low to high (LTH): blue, yellow, red. Edge (3,5) has the highest weight due to its visibility by the adversary. 
	In each subplot, the edge labels indicate the edge cost under the current conditions and show in brackets the desired number of robots, $a_e$, due to vulnerability. Overwatch opportunities are shown with a pink arrow from the overwatch node pointing to the edge that can be monitored. At each time step, the position of each robot scout is shown.}
	\label{fig:toy_problem_solution}
    \vspace*{-5mm}
\end{figure}

To demonstrate our results, we first consider the 
example in Fig.~\ref{fig:toy_problem_solution} which portrays a simple reconnaissance scenario. 
In Fig.~\ref{fig:toy_problem_solution_0}, the edge costs are shown as the maximum cost for one robot to traverse without any overwatch, teaming, or vulnerability cost reductions. 
To encourage teaming, on each edge, each additional robot reduces the cost by 1. Both directions of edges (1,3), (3,5), and (4,5) are considered vulnerable and at least four robots are desired; the cost reduction for each robot up to four is 10. Overwatch opportunities are indicated by the pink arrows. Overwatch can be provided from node 2 for both directions of edge (2,4) and node 3 for both directions of edge (4,5), which can reduce the cost by up to 20 or 60, respectively, when two robots are providing overwatch. 
Each additional robot providing overwatch after the first two would reduce the cost by 2. The dynamic costs incurred by each edge are updated in the graphs in Fig.~\ref{fig:toy_problem_solution} to reflect the teaming, vulnerability, and overwatch constructs being utilized. In this example we scale the time cost by 10 to encourage reaching the goal in minimal time. The total accumulated cost is tracked in the captions. 

In this solution, we see the robots break into three teams to work together for a team to safely traverse to the target node. The robots all move together on edge (1,2) to get the largest benefit from teaming. Two robots then remain at node 2 to provide overwatch to the four robots traversing edge (2,4) and the remaining four continue toward node 3 to be prepared to provide overwatch in the next step. The greatest benefits from overwatch at nodes 2 and 3 are with 2 robots. Four robots (rather than two) go to node 3 to also receive a larger teaming benefit on edge (2,3) and additional rewards for overwatch in the next step. It would have been equivalent cost for two of those robots to continue on with robots 4-7. In time step 3, robots 4-7 traverse a vulnerable edge, getting the largest reward for having four robots and enabling moving in a formation. In the last step, robots 4-7 reach the goal node. The cost does not increase since the robots moved to a node in this step and there is no longer time cost since there are not any robots remaining on edges. 

By solving our problem to optimality we are guaranteed to have minimized the total cost. However, this depends greatly on the weights/cost assigned to the problem and the prioritization of minimizing traversing cost versus minimizing time. 
For example, if we did not scale our time cost by 10, the resulting solution would keep more robots together to receive more teaming benefits for moving together and from additional overwatch cost reductions. As in Fig.~\ref{fig:toy_problem_solution_1}, all robots would go to node 2. Then the robots would split into only two teams: a team for overwatch that moves from node 2 to 3 and the team being monitored on edges (2,4) and (4,5). Ultimately requiring one additional time step. 

\subsection{Bounding Overwatch Example}

\begin{figure}[tbh]
	\vspace*{-3mm}
	\centering
	\includegraphics[trim={0cm 0.3cm 0cm 0.5cm},clip, width=\columnwidth]{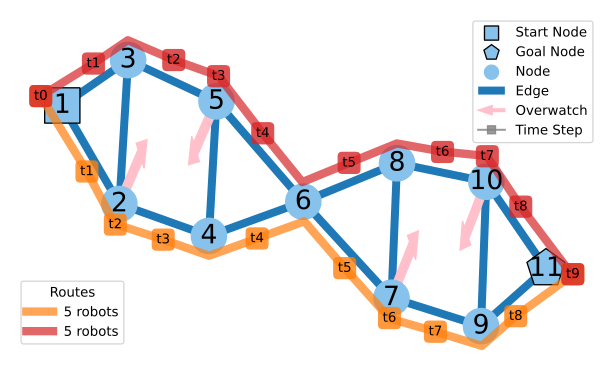}
	\caption{Example solution demonstrating bounding overwatch behavior as all robots move from node 1 to 11. Robot team routes are shown with each time step labeled. The overwatch opportunities, indicated by the pink arrows, are met at time steps 2, 3, 6, and 7. The two teams alternate providing overwatch as they move through the environment. 
	}
	\label{fig:problem4}
\end{figure}

As a demonstration of a larger graph that particularly lends itself to the bounding overwatch paradigm, Fig.~\ref{fig:problem4} shows a solution to our MIP problem for the graph shown in blue. We set the start point for all robots to be node 1 and the goal for all robots is node 11. The solution yields two robot teams alternating traversing and providing overwatch. Each team's route is visualized with time steps to enable correlating the teams' positions and the bounding overwatch behavior. 

\subsection{Real World Scenarios}

\begin{figure}[tbh]
	\vspace*{1.5mm}
	\centering
	\includegraphics[trim={1.9cm 0.4cm 1.9cm 0cm},clip, width=\columnwidth]{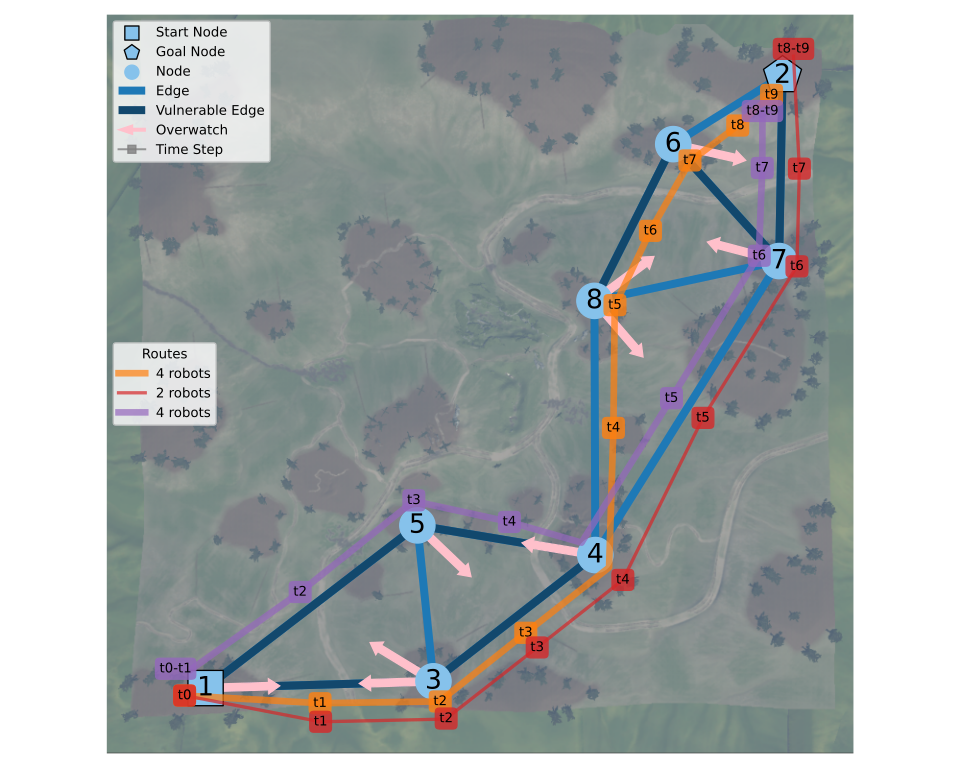}
	\caption{Aerial map of a meadow environment showing a sample scenario. Nodes are in regions of cover. Vulnerable edges due to crossing roads are shown in dark blue. The largest cost reductions on these edges come from having four robots. In the solution routes shown, robot teams split up and form new teams as they move through the terrain providing overwatch. 
	}
	\label{fig:built_graph}
	\vspace*{-3mm}
\end{figure}

In Figures~\ref{fig:built_graph} and \ref{fig:built_side_problem}, we demonstrate applying this approach to complex real world scenarios. We consider traversing between areas of forested cover in a high risk environment, using the meadow environment from \cite{NatureManufacture} as a representative example. We show results with all robots reaching a designated goal and a subset of robots reaching the goal. 

\subsubsection{Meadow Map 1} 

In Fig.~\ref{fig:built_graph}, we consider 10 robots starting at node 1 with a goal of all robots reaching node 2 within 10 time steps. We see three distinct paths emerge. Through time step 3, six robots maneuver together (orange and red teams) alternating providing overwatch with another team (purple) of four robots. At time step 4, two robots (red) remain at node 4 to provide overwatch for the purple team, while orange maneuvers toward the next overwatch position. This allows the purple and red teams to combine and proceed alternating overwatch with the orange team through the rest of the graph toward the goal node. 

\subsubsection{Meadow Map 2}

Fig.~\ref{fig:built_side_problem} is an example with further subdivision  
of the robots in the solution. Again we consider 10 robots starting at node 1. 
In this scenario, the goal is for at least one robot to reach node 2 within 12 time steps.
In this case, we see four teams emerge. One team of four (orange) that traverses through the map towards the goal, providing overwatch as needed, and three smaller supporting teams (red, purple, brown) providing overwatch for orange and each other. Since all robots do not need to reach the goal node, this example highlights the trade off between the cost of traversing and the benefit of overwatch. Teams will not continue towards the goal if the overall cost of their movement is not less than the benefit of the overwatch they would provide to a primary team moving towards the goal. This concept is consistent with expectations in an operational setting: the risk outweighs the reward. Ultimately, this is the objective of our method, to be able to determine strategic tactical maneuvers 
in complex environments. 

\subsection{Ablation Study}

\begin{figure}
	\centering
	\begin{subfigure}[t]{0.5\columnwidth}
		\footnotesize
		\centering
		\includegraphics[trim={0cm 0.5cm 0cm 0cm},clip,width=\textwidth]{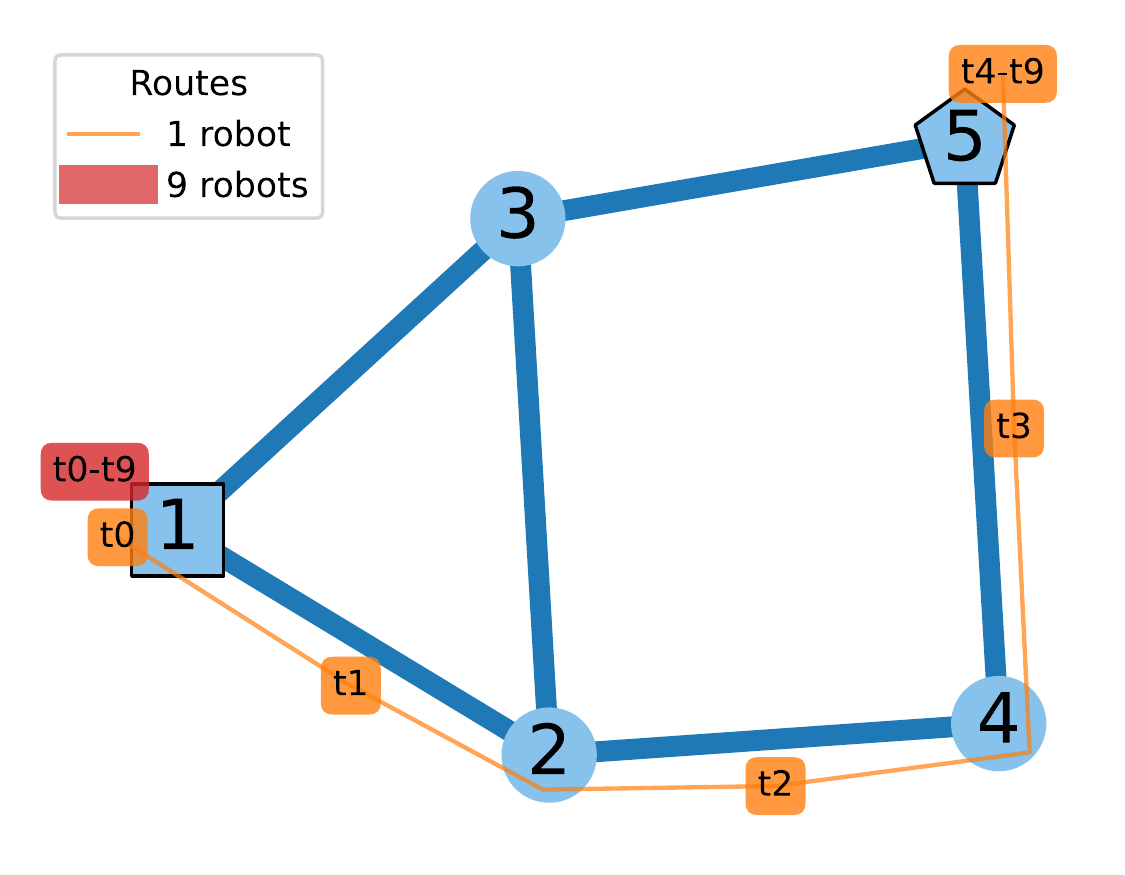}
		\scriptsize
        \caption{Without overwatch, vulnerability, \\and teaming}
		\label{fig:toy1_without_overwatch_vul_teaming}
	\end{subfigure}%
	\begin{subfigure}[t]{0.5\columnwidth}
		\footnotesize
		\centering
		\includegraphics[trim={0cm 0.5cm 0cm 0cm},clip,width=\textwidth]{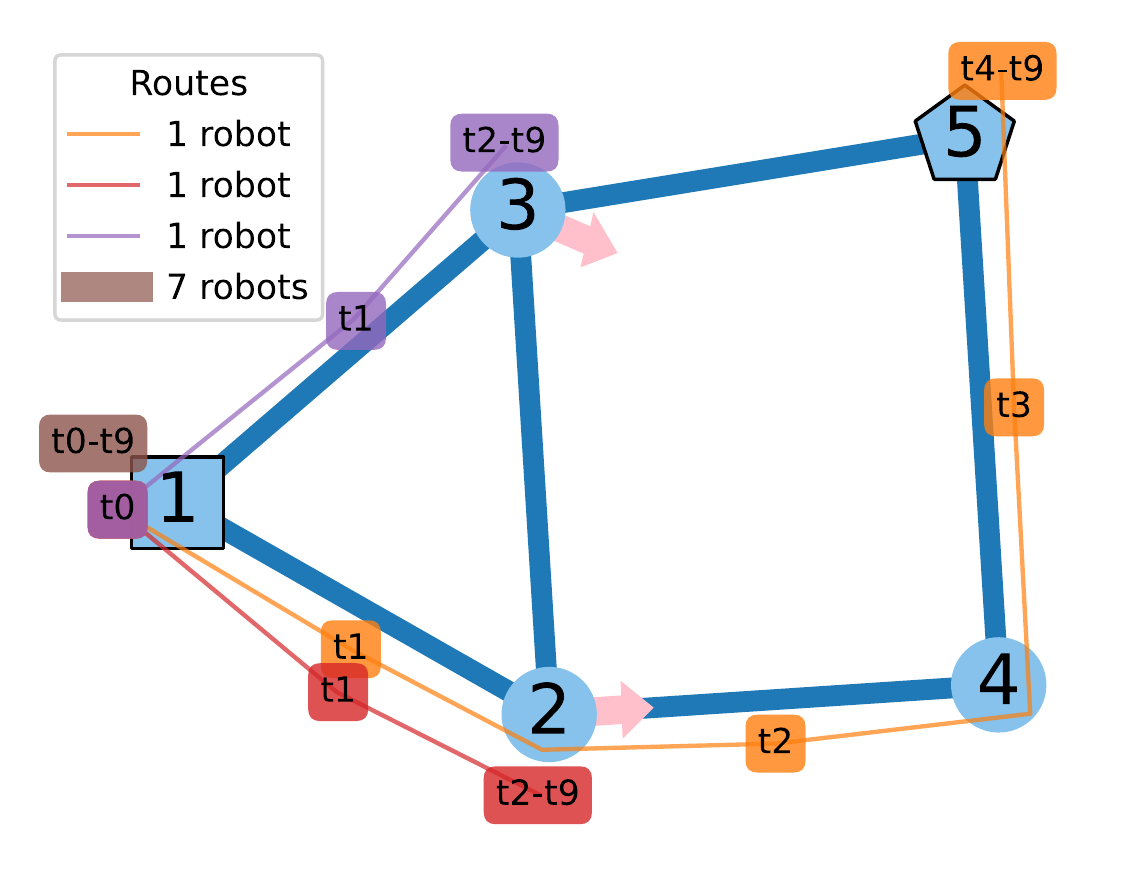}
		\scriptsize
        \caption{With overwatch and without vulnerability and teaming}
		\label{fig:toy1_without_vul_teaming}
	\end{subfigure}
	
	\begin{subfigure}[t]{0.5\columnwidth}
		\footnotesize
		\centering
		\includegraphics[trim={0cm 0.5cm 0cm 0cm},clip,width=\textwidth]{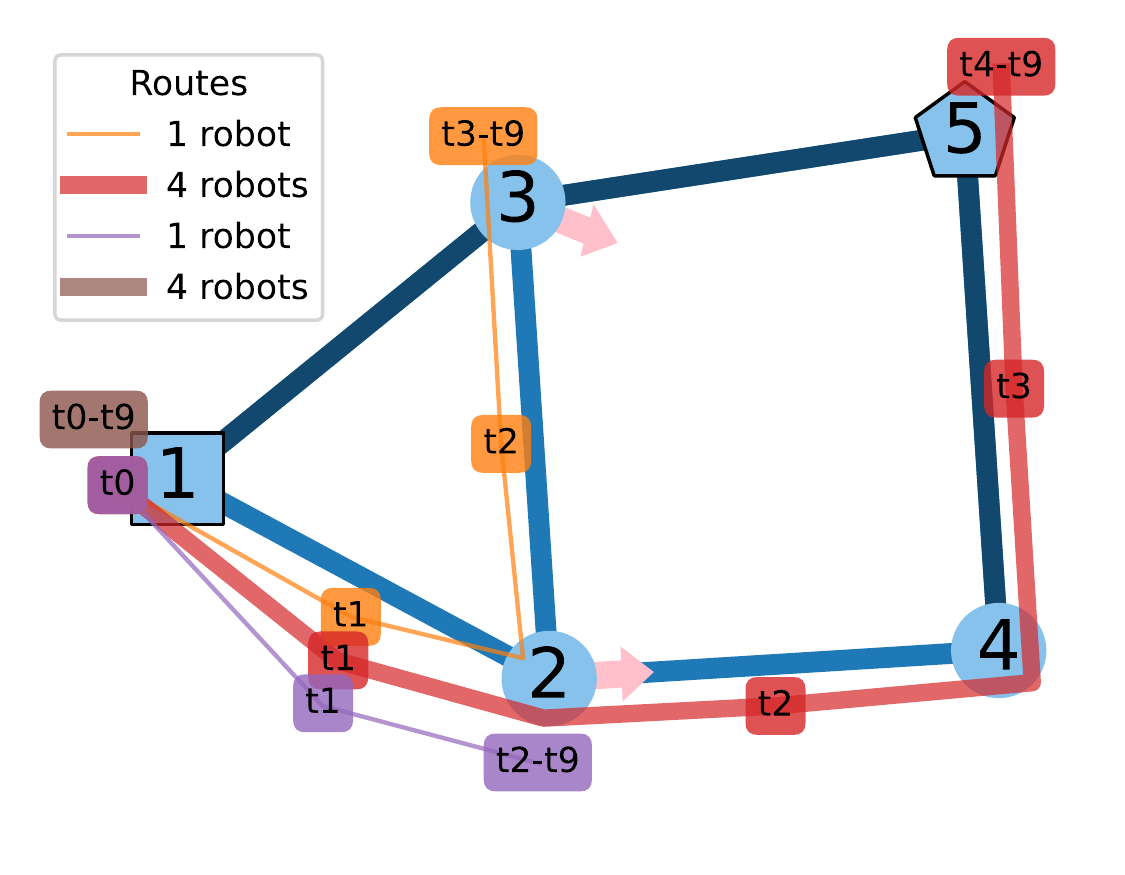}
		\scriptsize
        \caption{With overwatch and vulnerability \\and without teaming}
		\label{fig:toy1_without_teaming}
	\end{subfigure}%
	\begin{subfigure}[t]{0.5\columnwidth}
		\footnotesize
		\centering
		\includegraphics[trim={0cm 0.5cm 0cm 0cm},clip,width=\textwidth]{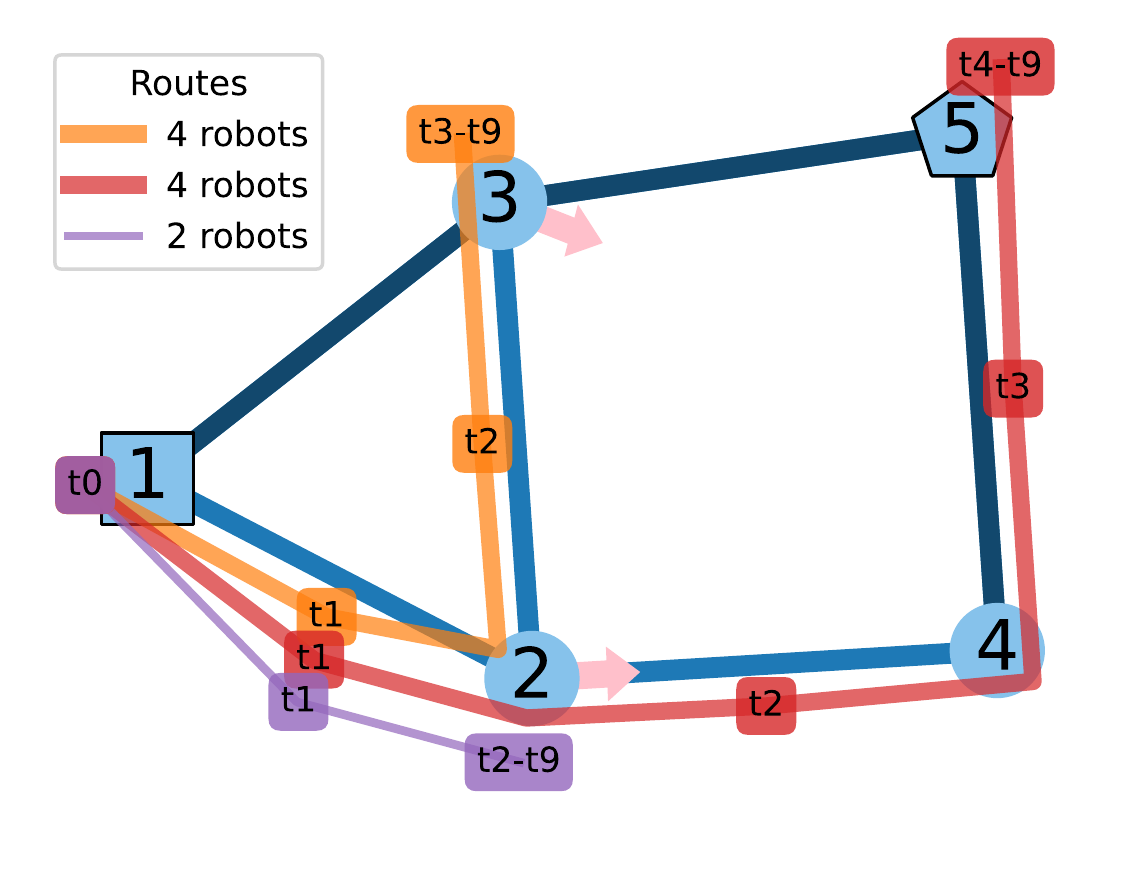}
		\scriptsize
        \caption{Full solution (with overwatch, vulnerability, and teaming)}
		\label{fig:toy_problem_solution_paths}
	\end{subfigure}%
	\vspace*{1.5mm}
	\begin{subfigure}[t]{\columnwidth}
		\centering
		\includegraphics[width=\textwidth]{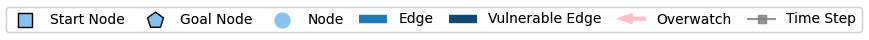}
	\end{subfigure}%
	\caption{Ablation study illustrating the impact of each of our main constructs: overwatch opportunities, edges with high vulnerability, and incentives for moving as a team. Robot team paths are shown in each subplot with components of our formulation incrementally added.}
	\label{fig:toy1_ablation}
    \vspace*{-3mm}
\end{figure}

We performed an ablation study to assess the impact of the 
constructs we label overwatch, vulnerability, and teaming in our algorithm. Fig.~\ref{fig:toy1_ablation} shows the solutions to our illustrative example when removing these components and incrementally adding them back in. 
The problem is setup as in Fig.~\ref{fig:toy_problem_solution}, with all robots starting at node 1 and an overall goal of at least one robot reaching node~5. 

When we remove the overwatch, vulnerability, and teaming constructs, in Fig.~\ref{fig:toy1_without_overwatch_vul_teaming}, the problem essentially becomes a shortest path problem. The weights in the graph are fixed and one robot takes the path with the least overall cost due to the edge weights and time. It would be an equivalent cost solution for any number of robots to take this path. 

In tactical maneuvers, overwatch allows minimizing detection and maximizing safety. When we add our formulation of overwatch opportunities in Fig.~\ref{fig:toy1_without_vul_teaming}, the edge weights now vary based on the overwatch opportunities being utilized and the optimal solution includes robots moving to the two overwatch positions while one robot traverses to the goal, overall making the operation safer. 

As an incentive to travel in a formation on edges that are particularly dangerous, or vulnerable, we add our vulnerability construct back in Fig.~\ref{fig:toy1_without_teaming}, making edges (1,3), (3,5), and (4,5) more costly to traverse alone. We see in the solution that the desired minimum of 4 robots traverse edge (4,5). The robot providing overwatch at node 3 moves with other robots on edge (1,2) and then alone on edge (2,3). This is ultimately the same cost as if four robots were to traverse the vulnerable edge (1,3). Both solutions are optimal. As in the previous cases, the extra robots stay at the start node. 

In a tactical scenario, more robots moving together would provide more data, potentially reducing the overall threat level. Additionally, the redundancy of having more robots yields a more robust solution. 
When we add our teaming construct back in,  Fig.~\ref{fig:toy_problem_solution_paths}, we see all robots moving for the first time, no longer leaving robots at the start node, since moving as a team provides cost reductions on each edge and greater cost reductions when providing overwatch. 
Ultimately, the overwatch, vulnerability, and teaming constructs yield practical plans for tactical maneuvers that minimize detection and maximize robustness and safe navigation.

\subsection{Computation Time}

\begin{table}[b]
    \vspace*{-2mm}
	\centering
	\caption{Example Graphs' Problem Size and Computation Time} 
	\label{tab:computation_time}
    \vspace*{-2mm}
	\begin{center}
		\renewcommand{\arraystretch}{1.3}
		\begin{tabular}{ c || c | c | c | c }
			& \textbf{Illustr.} & \textbf{Bounding} & \textbf{Map 1} & \textbf{Map 2} \\
			\hline \hline	
			\textbf{Locations, $n_L$} & 17 & 43 & 32 & 51 \\
			\textbf{Overwatch, $n_\scriptO$} & 4 & 8 & 18 & 32 \\
			\textbf{Time Horizon, $n_T$} & 10 & 10 & 10 & 12 \\
			\hline
			\textbf{Total Variables} & 460 & 1160 & 990 & 1872 \\
			\hline
			\textbf{Mean Solve Time (s)} & 0.081 &	0.100 &	0.249 &	3.702 \\	
		\end{tabular}
	\end{center}
    \vspace*{-2mm}
\end{table}

The common limitation of MIP is computation time. The high level planning we propose would need to be computed rapidly (on the order of magnitude of minutes) for robots to be able to act on the plan and re-plan as new data is gathered. Table~\ref{tab:computation_time} shows the parameters that affect the problem size for the graphs discussed in this paper (number of locations, number of overwatch opportunities, and time horizon), the total variables in our optimization problem due to those parameters, and the resulting solve time for our MIP problem in each case. 
The graphs are labeled Illustrative, Bounding, Map 1, and Map 2, which correspond to Figures~\ref{fig:toy_problem_solution} and \ref{fig:toy1_ablation}, Fig.~\ref{fig:problem4}, Fig.~\ref{fig:built_graph}, and Fig.~\ref{fig:built_side_problem}, respectively. 

We used the Gurobi optimizer \cite{GurobiOptimization2023} on an Intel® Core™ i7-10875H CPU @ 2.30GHz × 16 and averaged the solve times across 100 trials. In all cases the number of robots was 10. Varying the number of robots does not impact the total number of variables in our problem. 
We select suitable time horizons for each problem based on the size of the graph and location of the goal to assure the problem is feasible. 

These results show computation times in seconds for realistic operational scenarios. Additionally, the problem stays computationally tractable to   
solve complicated routes on large graphs for multi-robot teams.

\section{CONCLUSION}

In this work, we demonstrated expressing complicated scenarios compactly with dynamic topological graphs and MIP, significantly reducing the overall state space. 
We considered minimizing detectability in sample reconnaissance problems, introducing the concepts of overwatch, vulnerability, and teaming, and analyzed the trade-offs of these constructs. 
We show results solving complicated, real world scenarios in seconds, resulting in full tactical maneuvers for multi-robot teams with explicit collaboration. We have removed the dependence on the number of robots in our state space, allowing this approach to easily scale to large teams of robots. Additionally, for problems on larger graphs, we can plan with a receding horizon and as the environment changes we can continuously re-plan. 

By expressing this problem using MIP, the state space, costs, and constraints are fully comprehensible, and yield explainable results. We hypothesize that these results have the potential to generalize to larger classes of problems, such as search and rescue applications over treacherous terrain. We hope to extend this formulation for distributed planning and heterogeneous teams. Additionally, since we have formulated our problem convexly when integrality constraints are relaxed, we hope to find a tight relaxation to further decrease solve times. 





\section*{ACKNOWLEDGMENT}

We gratefully acknowledge the support of the Army Research Laboratory under grant W911NF-22-2-0241. 


\bibliographystyle{IEEEtran}
\bibliography{references.bib}

\end{document}